\documentclass[twoside,leqno,twocolumn]{article}

\usepackage[letterpaper]{geometry}

\usepackage{ltexpprt}
\usepackage{hyperref}
\usepackage{graphicx}
\usepackage{caption}
\usepackage{subcaption}
\usepackage{float}
\usepackage{amsmath}
\usepackage{amssymb}
\usepackage{comment}
\usepackage{color}

\begin{document}

\title{\Large Computing Steiner Trees using Graph Neural Networks}
%\author{Corey Gray\thanks{Society for Industrial and Applied Mathematics.}
%\and Tricia Manning\thanks{Society for Industrial and Applied Mathematics.}}
\author{Reyan Ahmed \and Md Asadullah Turja \and Faryad Darabi Sahneh \and Mithun Ghosh \and Keaton Hamm \and Stephen Kobourov}

\date{}

\maketitle

% Copyright Statement
% When submitting your final paper to a SIAM proceedings, it is requested that you include
% the appropriate copyright in the footer of the paper.  The copyright added should be
% consistent with the copyright selected on the copyright form submitted with the paper.
% Please note that "20XX" should be changed to the year of the meeting.

% Default Copyright Statement
%\fancyfoot[R]{\scriptsize{Copyright \textcopyright\ 20XX by SIAM\\
%Unauthorized reproduction of this article is prohibited}}

% Depending on which copyright you agree to when you sign the copyright form, the copyright
% can be changed to one of the following after commenting out the default copyright statement
% above.

%\fancyfoot[R]{\scriptsize{Copyright \textcopyright\ 20XX\\
%Copyright for this paper is retained by authors}}

%\fancyfoot[R]{\scriptsize{Copyright \textcopyright\ 20XX\\
%Copyright retained by principal author's organization}}

%\pagenumbering{arabic}
%\setcounter{page}{1}%Leave this line commented out.

\begin{abstract} \small\baselineskip=9pt Graph neural networks have been successful in many learning problems and real-world applications. A recent line of research explores the power of graph neural networks to solve combinatorial and graph algorithmic problems such as subgraph isomorphism, detecting cliques, and the traveling salesman problem. However, many NP-complete problems are as of yet unexplored using this method. In this paper, we tackle the Steiner Tree Problem. We employ four learning frameworks to compute low cost Steiner trees: feed-forward neural networks, graph neural networks, graph convolutional networks, and a graph attention model. We use these frameworks in two fundamentally different ways: 1) to train the models to learn the actual Steiner tree nodes, 2) to train the model to learn good Steiner point candidates to be connected to the constructed tree using a shortest path in a greedy fashion. We illustrate the robustness of our heuristics on several random graph generation models as well as the SteinLib data library. Our finding suggests that the out-of-the-box application of GNN methods does worse than the classic 2-approximation method. However, when combined with a greedy shortest path construction, it even does slightly better than the 2-approximation algorithm. This result sheds light on the fundamental capabilities and limitations of graph learning techniques on classical NP-complete problems. 
\end{abstract}

\section{Introduction}
Graphs are one of the most flexible data structures for capturing relational information. Classical machine learning models such as neural networks or recurrent neural networks are not designed to handle graph input or output directly.  
%The graph is the most flexible data structure to relate nodes with other nodes using links. Traditional neural networks are not designed to learn graphs. Even recurrent neural networks are only applicable to sequential information. 
To learn a flexible data structure like a graph, Gori et al.~\cite{gori2005new} introduced the concept of a graph neural network (GNN). They represented the GNN as a recursive neural network where nodes are treated as state vectors, and the relationship among these nodes is quantified by the edges. Scarselli et al.~\cite{scarselli2008graph} extended the notion of unfolding equivalence that leads to the transformation of the approximation property of feed-forward networks (Scarselli and Tsoi~\cite{scarselli1998universal}) to GNNs.

Many combinatorial problems arise naturally in various applications, many of which are known to be NP-complete. In 1972, the seminal paper~\cite{karp1972reducibility} provided a list of NP-complete problems, many of which are graph-based. GNNs offer an alternative to traditional heuristic approximation algorithms; indeed the initial GNN model~\cite{scarselli2008graph} was used to approximate solutions to two classical graph-based problems: subgraph isomorphism and clique detection.

An interesting line of research is to explore the capabilities and limitations of graph neural networks as alternative approaches to existing heuristics and approximation algorithms to theoretical graph problems. This can open up modern methods to solve such problems at scale in practice, and also it can shed light on the fundamental capacity of such learning methods.

\subsection{Background}
Graph neural networks have been a very active area of research during the past few years. To name a few, Lei et al.~\cite{lei2017deriving} introduced recurrent neural operations for graphs with their associated kernel spaces. The notion of studying graph neural models as Message Passing Neural Networks is discussed by Gilmer et al.~\cite{gilmer2017neural}. Garg et al.~\cite{garg2020generalization} generalized the standard message passing GNNs that rely on the local graph structure, proposing a novel framework for the GNN through graph-theoretic formalism which provided insights into the design of more effective GNNs. 

Mishra et al.~\cite{mishra2020node} demonstrated state-of-the-art spatial GNNs operations with theoretical tools. They proposed the node masking concept for better generalization and scaling in both transductive and inductive settings.
Most recent works on learning the graph structure data focus on the distributed representation of the substructure of the graph such as nodes and subgraphs. But for the tasks related to graph clustering or graph classification, models require the entire graph as fixed-length feature vectors.  Graph kernels remain as one of the most effective ways to obtain these vectors by using some handcrafted features such as shortest paths or graphlets, but this technique can also result in poor generalization. Narayanan et al.~\cite{narayanan2017graph2vec} proposed a neural embedding formulation named graph2vec to learn data-driven distributed representations of arbitrary graphs.
The embeddings of graph2vec are learned in an unsupervised manner and are task agnostic.
Different explainer models have been also proposed to understand the reason for decision-making by the models~\cite{dai2020framework,ying2019gnnexplainer}.
%To learn a flexible data structure like a graph, Scarselli et al.~\cite{scarselli2008graph} proposed the graph neural network model to learn graphs. Later, their model has been extended and

\subsection{Related Works}

Graph neural networks have been widely used in many areas including physical systems~\cite{battaglia2016interaction,sanchez2018graph}, protein-protein interaction networks~\cite{fout2017protein}, social science~\cite{hamilton2017inductive,kipf2016semi}, and knowledge graphs~\cite{hamaguchi2017knowledge}. For more information on graph neural networks, see the survey~\cite{zhou2018graph}.

  %Today we know many other NP-hard problems. A large number of such problems are graph-based. Hence, the graph neural network is a suitable tool to handle such problems. Indeed, the initial GNN model~\cite{scarselli2008graph} discussed two classical graph-based problems: subgraph isomorphism and clique detection. 

Subsequently, other methods have been proposed to solve other combinatorial problems; for example, Khalil et al.~\cite{khalil2017learning} proposed a framework to compute minimum vertex cover, maximum cut, and traveling salesman problems. The survey~\cite{vesselinova2020learning} discusses different combinatorial problems that have been approached using graph neural networks or related methods. %Recently, Prates~\cite{prates2019learning} et al. proposed a GNN model to solve the traveling salesman problem. %We extend this model to solve the Steiner tree problem.
%{\color{red}  EXPAND BACKGROUND WORK ON GNNS FOR ALRORITHMS. CITE THE GRAPH COLORING PROBLEM, AND SEVERAL OTHERS IN THE REVIEW PAPER YOU MENTION ABOVE.}
Bello et al.~\cite{bello2016neural} studied machine learning in general and deep reinforcement learning in particular for the planar traveling salesman problem (TSP). Kool et al.~\cite{kool2018attention} also studied the TSP problem to make progress in learning heuristics that can be applied to a broad scope of different practical problems. Prates~\cite{prates2019learning} et al. proposed a message-passing GNN model to predict the decision TSP problem. Lemos et al.~\cite{lemos2019graph} show that GNNs can be trained to solve fundamental combinatorial optimization challenges such as the
graph coloring problem. Different neural network models have been proposed to solve other combinatorial problems such as satisfiability (SAT) problem~\cite{selsam2018learning,li2018combinatorial}, vertex cover~\cite{li2018combinatorial,dai2017learning}, independent set~\cite{li2018combinatorial} etc. To the best of our knowledge, there does not exist any work that considered the Steiner tree problem.
\subsection{Problem Statement}
The Steiner Tree Problem is another classical problem mentioned in Karp's initial NP-complete problem list~\cite{karp1972reducibility}. In this problem, we are given a weighted graph $G = (V, E)$ and a set of terminals $T \subset V$, and the objective is to compute a minimum cost tree that spans $T$. Several approximation algorithms have been proposed for this problem including a classical 2-approximation algorithm that first computes the metric closure of $G$ on $T$ and then returns the minimum spanning tree~\cite{agrawal1995trees}.

In this paper we are interested in the following question: can graph neural networks provide good candidate solutions to the Steiner Tree Problem?

\subsection{Summary of Contributions}
%{\color{red} TODO: PLEASE REWORD TO MAKE IT DIFFERENT FROM ABSTRACT}

In this paper, we tackle the Steiner Tree Problem from a graph neural network perspective. Specifically,
\begin{itemize}
    \item %We develop four different neural network models to compute the Steiner trees.
    We propose four different models to predict solution nodes for the Steiner tree problem.
    
    \item %We use these models in two fundamentally different ways: 1) to train the models to learn the actual Steiner tree nodes, 2) to train the model to learn good Steiner points candidate to be connected to the constructed tree using a shortest path in a greedy fashion.
    We propose two heuristics to compute low cost Steiner trees from these models. In one heuristic we use the model's predictions to construct a connected induced graph. We then compute the spanning tree of that induced graph. In the other heuristic, we predict good Steiner nodes and connect them to the constructed tree using a greedy shortest path method.
    
    \item %We illustrate the robustness of our heuristics on several random graph generation models as well as the SteinLib data library. 
    We generate forty thousand Steiner tree instances using different random graph generators. We compute the exact solution of these instances, and train and test the models on this dataset. We also test the models on the SteinLib data library which is outside of the training set.
\end{itemize}

Our finding suggests that out-of-the-box application of GNN methods does worse that the classic 2-approximation method. However, when combined with a greedy shortest path construction, it even does slightly better than the 2-approximation algorithm.

\section{Neural network-based models}

To begin, we lay out the different learning models that will be tested in Section \ref{sec:experiment}. The first model is not a graph learning model, but rather a feedforward model that turns out to not be well-suited for graph problems. However, we include it to get a baseline for the performance of other models.

\subsection{Feedforward model}
In the Steiner tree problem, given an input graph $G$, we want to determine which nodes are present in the Steiner tree. The problem can be represented by a function $f(G)$ that outputs a binary vector $Y$. The value of $Y[i]$ is 1 if and only if the $i$-th node is present in the solution. One can use feedforward neural networks to approximate this function. A feedforward network defines a mapping $y=f(x, \theta)$ and learns the value of the parameters $\theta$ that result in the best function approximation. The input of a feedforward network is a vector, and one can represent a graph by a binary vector $x$ where each element corresponds to a pair of nodes. An entry of $x$ is 1 if and only if there is an edge between the corresponding pair of nodes. In this representation, we need to assume that the number of node pairs or number of nodes will not exceed an upper bound. Indeed, we build the model for the graph's size of the upper bound and treat any other smaller graph as a large graph with auxiliary, isolated non-terminal nodes.

\subsection{Graph neural network model}
A GNN model, in a transductive-inductive framework was proposed by~\cite{scarselli2008graph} 
%{\color{red}[1] -- this is a hardcoded reference and is probably incorrect --} 
using the TensorFlow platform. %Whereas an edge consists of a pair of nodes for example, $e=(m,n),e \in E$ and $m,n \in V$. The order of the pair of nodes in the edge is important when we need to consider the asymmetric relationship. For this reason, edges are also called arcs for the directed graph. 
The GNN model is capable of handling both directed and undirected graphs. In this framework, both edges and nodes have attributes that are called labels (typically vectors in some parameter space). Examples of node labels are average colors, area, and shape factors, and examples of edge features are barycenters of two adjacent regions of a triangulation, or distance between nodes. Each node also has a state, and a computational process named diffusion updates the states. The input graph topology drives the computation. The diffusion mechanism forms the computation schema which updates the state vector at each node as a function of node labels, neighboring edge labels, and states of the neighboring nodes.
The information relevant to each task will be summarized by the state for each node. The node states are used to compute the class or target properties. %{\color{red} [K]: This whole part is unclear.}

Let the state and output at node $n\in V$ be $x_n \in \mathbb{R}^s$ and $o_n \in \mathbb{R}^m$, respectively. Also, let $d_w$ represents the state transition function which ultimately drives the diffusion process and $f_w$ denotes the output function. The diffusion process can be described by the following equations:

\begin{equation}
x_n = \sum_{(n, v) \in E} d_w(l_n, l_{(n, v)}, x_v, l_v),
\end{equation}

\begin{equation}
    o_n = f_w(x_n, l_n).
\end{equation}

Where $l_n \in \mathbb{R}^q$ and $l_{(n, v)} \in \mathbb{R}^p$ are the labels attached to $n$ and $(n,v)$, respectively. The solution of the Jacobi iterative procedure can be written as

\begin{equation}\label{eqn:unfolding}
    x_n (t+1)=\sum_{(n,v) \in E} d_w (l_n,l_{(n,v)}), x_v (t),l_v).
\end{equation}

\begin{figure*}[t]
  \centering
    \centering
    \includegraphics[width=2\columnwidth]{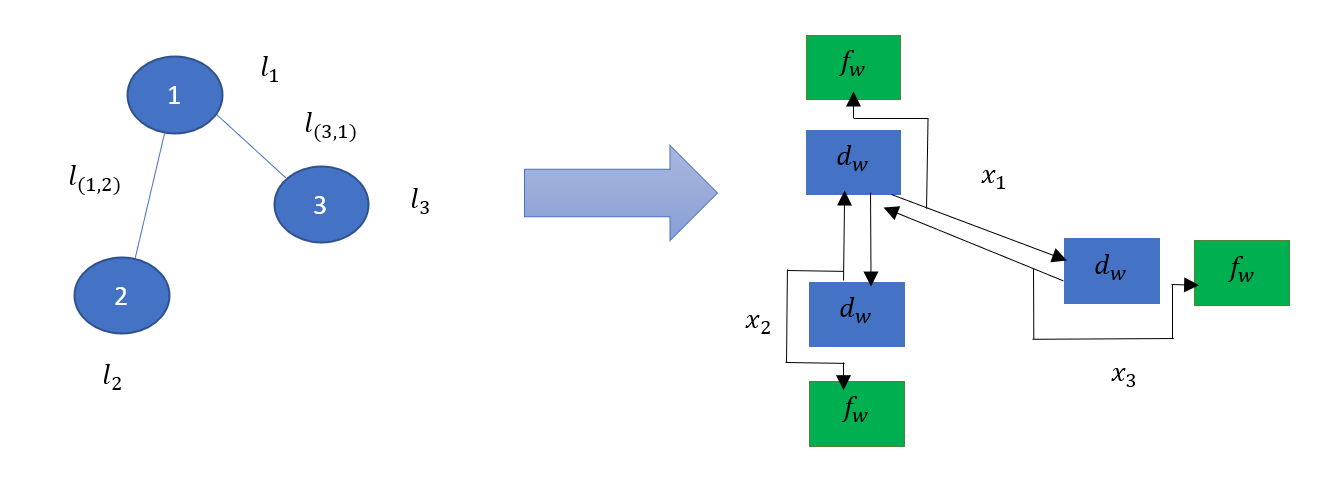}
    \caption{Encoding network framework corresponding to a given input graph (from left to right). For each node, $d_w$ and $f_w$ processing units are replicated and connected following the graph topology.}
    \label{fig:unfolding}
\end{figure*}

Equation~\eqref{eqn:unfolding} implements the diffusion process for the state computation. Simple multilayer perceptrons (MLPs) with a unique hidden layer define the functions $d_w$ and $f_w$ in this framework. The unfolding of the encoding network (Figure~\ref{fig:unfolding}) is demonstrated by Equation~\eqref{eqn:unfolding} by calculating the $d_w$ and $f_w$ for each node. Each node represents a replica of perceptron realizing $d_w$ whereas each unit $x_n(t)$ represents the state at time $t$. The next time state at time $t+1$ is computed by the stored states in all nodes at time $t$. When the state computation converges, $f_w$ is applied to each node to compute the output. The error backpropagation is performed on the unfolding network for the gradient computation. The weights of the network are adopted such that it reduces the error between the output of the network and the expected targets on the supervised node sets during the training. This procedure is discussed further in Section \ref{sec:training}.

\subsection{Graph Convolution Network} \label{sec:gcn}
Graph Convolution Networks (GCNs) generalize the notion of convolution from grid data to graph data. Essentially, a GCN learns a desired set of node features from it's own features and the neighboring nodes features. Unlike GNNs, GCNs stack multiple convolution layers to learn higher order node representations. We can consider the convolution operation as the multiplication of a graph signal $x \in \mathbb{R}^n$ with a filter $g_\theta = \textnormal{diag}(\theta)$ in the Fourier domain, i.e.,
\begin{equation} \label{eq:spgcn}
    g_\theta * x = U g_\theta U^T x.
\end{equation}

Here $U$ is the matrix of eigenvectors of the symmetric normalized graph Laplacian $L = I_n - D^{-\frac{1}{2}} A D^{-\frac{1}{2}} = U \Delta U^T$, $U^Tx$ is a graph Fourier transform of $x$, and $\Delta$ is a diagonal matrix containing the eigenvalues of $L$ (see, e.g., \cite{shuman2016vertex} for more details on graph Fourier transforms). By comparing $L = U \Delta U^T$ with the graph convolution operator in Equation~\ref{eq:spgcn}, we can interpret $g_\theta$ as function of $\Delta$, i.e, $g_\theta(\Delta)$ 
%{\color{red}[K]: This isn't clear to me}. 
Computing Equation~\eqref{eq:spgcn} can prohibitively expensive because computing the spectral decomposition of the graph Laplacian is computationally expensive for large graphs ($O(n^3)$ flops). In this regard, \cite{HAMMOND2011129} suggested an approximation of $g_\theta(\Delta)$ by means of Chebyshev polynomials $T_k(x)$ up to $K^{th}$ order:
\begin{equation} \label{eq:chebply}
    g_{\theta'}(\Delta) = \sum_{k=0}^{K} \theta'_k T_k(\hat{\Delta}),
\end{equation}
where $\hat{\Delta} = \frac{2}{\lambda_{max}}\Delta - I_N$. $\lambda_{max}$ denotes the largest eigenvalue of $L$ and $\theta' \in R^n$ is a vector of Chebyshev coefficients (More details about Chebyshev polynomial can be found at \cite{HAMMOND2011129}).
%{\color{red}[K]: Not clear to me what $\theta'$ is. There is no $\hat{\Delta}$ in Equation \eqref{eq:chebply}. $T_k(x)$ makes sense, but in the equation you have $T_k(\hat{L})$ which doesn't.}.
Note that $g_{\theta'}$ is a $K$-localized filter, which means it depends only on nodes that are at most $K$ steps away from the source node, since it is a $K^{th}$ order polynomial. The complexity of computing \eqref{eq:chebply} is linear in the the number of edges in the graph. In \cite{kipf2016semi}, the authors built a layer-wise convolutional neural network of \eqref{eq:chebply} assuming $K=1$. Each layer is followed by a pointwise nonlinearity, i.e., a function that is nonlinear w.r.t $L$. In our work, we use this version of GCN due to it's simplicity and ability to learn a rich class of convolutional filters by stacking multiple layers. Under this approximation, Equation~\eqref{eq:chebply} simplifies to:
\begin{equation}
    g_\theta * x \approx \theta'_0x + \theta'_1(L - I_N)x = \theta'_0x - \theta'_1 D^{-\frac{1}{2}}AD^{-\frac{1}{2}}x,
\end{equation}

\textbf{Our GCN model architecture}: We stacked two-layers of GCN followed by a multi-layer perceptron (MLP) with two hidden layers for our Steiner node predictor network (Figure~\ref{fig:gcn_gat}). The size of the hidden layers are 128 for both the GCN and MLP layers. Finally, we minimized cross-entropy loss to train our network for 500 epochs with a learning rate of 1e-3.

\subsection{Graph Attention Network}
The GCN in Section~\ref{sec:gcn} is capable of learning a wide range of kernels. However, the learned filters depend on the eigenbasis of the graph Laplacian, which hinders it's generalization power. For this reason, \cite{velickovic2018graph} proposed Graph Attention Networks (GATs) by stacking layers in which nodes are able to figure out which neighboring nodes are important to compute it's updated node features. GATs enable specifying different weights to different nodes in a neighborhood without depending on knowing the graph structure up front. The attention module in a GAT is a single layer feed-forward network followed by a LeakyReLU nonlinearity. Initially, a linear transformation is applied at each node feature; then a shared attentional mechanism $a$ computes the attention coefficient:
\begin{equation}
    e_{ij} = a(W x_i, W x_j),
\end{equation}
were $W$ is the learned weight matrix applied to the features $x_i$ and $x_j$ of nodes $i$, and $j$, respectively. The quantity $e_{ij}$ is called the attention score as it arises from the attention module $a$.

\begin{figure}[t]
    \centering
    \includegraphics[width=\columnwidth]{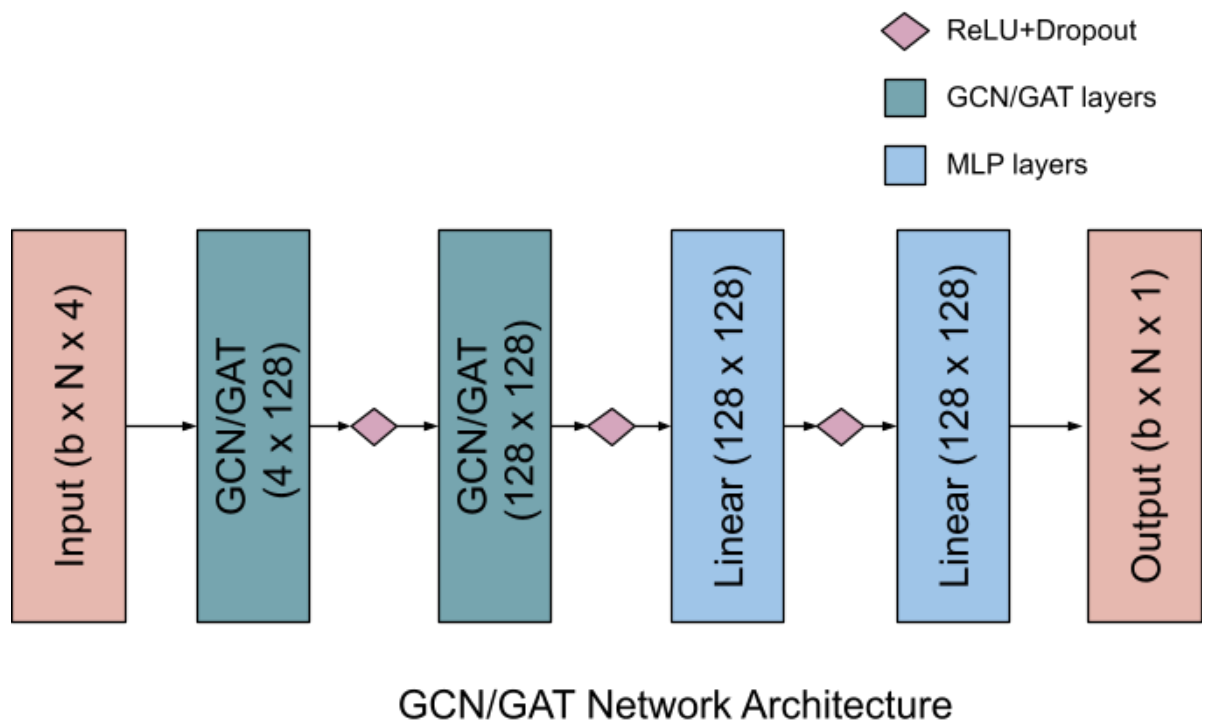}
    \caption{Network Architecture for the GCN and GAT models. Each of them consists of two layers of GCN/GAT layers followed by a multilayer perceptron (MLP). The hidden layers of both GCN/GAT and MLP are followed by pointwise nonlinear ReLU and Dropout layers.}
    \label{fig:gcn_gat}
\end{figure}

In our GAT model, we used a similar architecture as GCN by replacing GCN modules with GAT (Figure~\ref{fig:gcn_gat}). We also used the same hyperparameters and loss function as for the GCN.

\section{Two frameworks of using the learning models}

\subsection{The learning-only black-box method}
In this method, we just show the model several STP instances and its optimal solution and see if the model can correctly learn to identify nodes belonging to Steiner trees.

In this framework, the neural network models provide likelihood scores that each node is in the Steiner tree solution. Thus, we must determine which edges should be included to make the node set into a candidate solution for the Steiner tree. We use two heuristics to do this here.  To begin, we first add the terminal nodes to the solution if they are not already present, and then compute the induced graph. We keep adding the rest of the nodes one-by-one by decreasing order of the likelihood score and forming the induced graph until we have a connected induced graph. In the last step, we compute a minimum spanning tree of the induced graph and prune non-terminal nodes that have degree 1. Checking the connectivity of the induced graph can be computed in $\Tilde{O}(m)$, where $m$ is the number of edges~\cite{cormen2009introduction}. The running time of computing the minimum spanning tree is also $\Tilde{O}(m)$~\cite{cormen2009introduction}.

\subsection{A learning-assisted heuristic: combining learning and algorithms}\label{sec:heuristic}
In this approach, we compare the simple heuristic with a classical $2$-approximation algorithm. In this approximation algorithm, given an input graph $G = (V, E)$ and a set of terminals $T$, we first compute a complete graph $G' = (T, E')$. The edge weights of $G'$ are set equal to the shortest path distance between the terminal endpoints. The graph $G'$ is called the metric closure of $G$. The minimum spanning tree of the metric closure provides a Steiner tree of $G$ on terminals $T$. %The running time of the algorithm is dominated by the computation of metric closure, which is $\Tilde{O}(|T|m)$ 
%The Steiner tree problem is well-studied and there exists different approximation algorithms, see~\cite{hauptmann2013compendium,promel2012steiner,winter1987steiner}. However, the $2$-approximation algorithm has a straightforward implementation and performs well in practice. We also use this algorithm as a baseline algorithm for our experiments. {\color{red}[K]: This paragraph should go elsewhere; probably in the experiment section.}

The simple heuristic provides a valid Steiner tree; however, in our numerical experiments, we find that the approximation ratio for this heuristic (ratio of the cost of the returned solution and the cost of the optimal solution) is often larger than that of the $2$-approximation algorithm discussed in Section~\ref{sec:experiment}.  Therefore, we also tested another heuristic for forming the Steiner tree from the learning model output as follows.

\begin{figure}
    \centering
    \includegraphics[width=.5\linewidth]{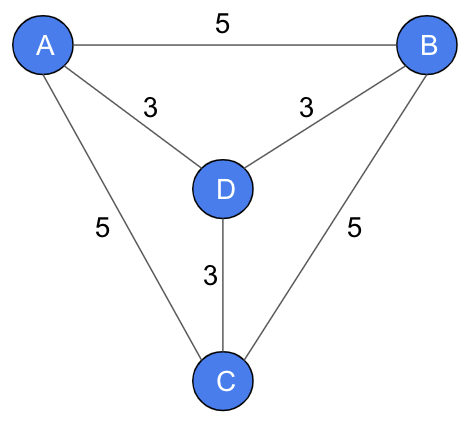}
    \caption{Example graph for the heuristic to compute the Steiner tree.}
    \label{fig:heuristic}
\end{figure}

In Figure~\ref{fig:heuristic}, A, B and C are the terminal nodes whereas D is not. The $2$-approximation algorithm will first construct the metric closure by computing the shortest paths of all pair terminals. Note that the non-terminal vertex D does not appear in any shortest path. For example, the shortest path length from A to B is 5, and this is from the direct edge connecting the terminals. Hence, the non-terminal node will not be present in the $2$-approximation. Without loss of generality, the $2$-approximation algorithm will choose the A-C-B path with a total cost of 10. Even though the path contains all the terminal nodes, this is not the optimal solution in terms of cost. We update the heuristic with a myopic decision such that the heuristic chooses the optimal cost that may also include nodes that are not terminal. %provided that all the terminal nodes will be included in the solution. 
For example, the edges AD, DB, and DC form the optimal solution, it contains all the terminal nodes with the minimum cost of 9. The $2$-approximation algorithm does not add any node that does not belong to a shortest path between two terminal nodes. We use our neural network models to predict nodes that may not belong to any shortest path, but whose inclusion improves the solution. We include these predicted nodes as terminals and then compute the $2$-approximation algorithm. Each time we include a new node, we run the $2$-approximation. How many new nodes will be added can be considered as a parameter of the heuristic. For our implementation, we keep adding nodes until the induced graph of the set of current nodes is connected. %{\color{red}The figure is not well-explained at all here.} %The running time of the algorithm is $\Tilde{O}(km)$, where $k$ is number of nodes we considered as terminal nodes.

\section{Model setup and training}\label{sec:training}

In order to train the models, one has to provide training data consisting of input graphs $G = (V, E)$, edge weights $W:E \rightarrow \mathbb{R}^+$, and terminals $T \subseteq V$. Given $G, W, T$, our goal is to produce a binary label for each vertex in $G$, such that label 1 indicates that a vertex is in the Steiner tree and label 0 indicates that it is not. The model is trained with Stochastic Gradient Descent (SGD) using the ADAM optimizer~\cite{kingma2014adam} to minimize the binary cross-entropy loss between the models' prediction and the ground-truth (a boolean vector in $\{0, 1\}^{|V|}$ indicating whether a node for the Steiner tree problem is in the solution or not) for each training sample.

\subsection{Data generation}
We produce training instances using several different random graph generation models:  Erd\H{o}s--R\'{e}nyi (ER)~\cite{erdos1959random}, Watts--Strogatz (WS)~\cite{watts1998collective}, Barab\'{a}si--Albert (BA)~\cite{barabasi1999emergence}, and random geometric (GE)~\cite{penrose2003random} graphs. For (ER), there is an edge selection probability $p$, which we set to be at least $\frac{2\ln n}{n}$ to ensure that the generated graphs are connected with high probability. In the (WS) model, we initially create a ring lattice of constant degree $K=6$. We then rewire each edge with probability $0 \leq p \leq 1$ while avoiding self-loops or duplicate edges. For our experiments we use $K=6$ and $p=0.2$. In the (BA) model, a new node is connected to $m=5$ existing nodes.
In the random geometric graph model, we uniformly select $n$ points from the Euclidean cube, and connect nodes whose Euclidean distance is not larger than a threshold $r_c$, which we choose to be $\sqrt{\frac{(1+\epsilon)\ln n}{\pi n}}$ for some $\epsilon>0$ to ensure the graph is connected with high probability.
%We generate a set of small graphs ($10 \le n \le 40$) and a set of large graphs ($50 \le n \le 500$). We only compute the exact solutions for the small graphs since the ILP has an exponential running time.

The Steiner tree problem is NP-complete even if the input graph is unweighted~\cite{garey1979computers}. We generate both unweighted and weighted Steiner tree instances using the random generators described above. The number of nodes of these instances is in $\{10, 20, \cdots, 200\}$. For the number of terminals, we use two distributions. In the first distribution, the percentage of the number of terminals with respect to the total number of terminals is in $\{20\%, 40\%, 60\%, 80\%\}$. In the second distribution the percentage is in $\{3\%, 6\%, \cdots, 18\%\}$. These two cases are considered to determine the behavior of the learning models on large and small terminal sets (compared with the overall graph size). %Because for a larger number of terminals the problem is similar to the minimum spanning tree problem, which is a polynomially solvable problem.

%ILP
%Python
%Cplex
%Hardware
%total number of instances
%SteinLib
\subsection{Computing optimal solutions}
We have generated around 40,000 Steiner tree instances. For each of these instances, we have computed the exact solution via a known flow-based integer linear program (ILP)~\cite{ahmed2019multi}. While there are various ILPs for solving the Steiner tree problem, our choice of this ILP is due to its fast runtime compared with others.  %{\color{red}There are different integer programs for the Steiner tree problem. The cut-based approach considers all possible combinations of partitions of terminals and ensures that there is an edge between that partition. This ILP is simple but introduces an exponential number of constraints. Another type of ILP considers an arbitrary terminal as a root and sends a flow to the rest of the terminals from the root. The flow-based ILP runs relatively faster and we used this ILP to compute exact solutions. More details of different types of ILP can be found here~\cite{ahmed2019multi}.}

We used CPLEX 12.6.2 as an ILP solver in a high-performance computer for all experiments (Lenovo NeXtScale nx360 M5 system with 400 nodes). 
Each node has 192 GB of memory. We have used Python for implementing the algorithms and spanner constructions. Since we have run the experiment on a couple of thousand instances, we run the solver for four hours to solve each instance. The exact solution of each instance of ER, WS, and BA random graph generator was able to finish in four hours. For GE instances, 99.17\% of the instances were able to finish in four hours. %{\color{red}[K]: What do you mean run the solver for 4 hours? Was this just how long it took to run all the instances, or what?}

While random graphs can sometimes be ideal, and thus the Steiner tree problem on them may be easily solved, the SteinLib library~\cite{KMV00} provides a catalog of hard graph instances for solving the Steiner tree problem. For thorough comparison, we perform experiments on instances from two subsets of SteinLib: \href{http://steinlib.zib.de/showset.php?I080}{I080} and \href{http://steinlib.zib.de/showset.php?I160}{I160}.

\subsection{Model architectures}
For the feedforward model, we have used two hidden layers each having 100 neurons with a ReLU activation function. For the output layer, we have used the sigmoid activation function. For the graph neural network model, we set the size of the state dimension equal to 5. For the multi-layer perceptrons representing $d_w$ and $f_w$ we have used one hidden layer of size 40 with the tanh activation function. For the graph convolutional network and attention model, we have used two hidden layers of size 128. For the GCN model, we have used the ReLU activation function. For the GAT model, we have used the ELU activation function.

\subsection{Feature selection}

We have used different properties of input instances as features to train the neural networks. We provide a list of these features here:

\begin{enumerate}
    \item \textbf{Shortest paths:} For every pair of vertices, we compute the shortest path. We add the shortest path distance as a feature.
    \item \textbf{Vertex degree:} For every vertex we use the number of adjacent vertices (degree) as a feature. We denote the degree of vertex $u$ by $deg(u)$.
    \item \textbf{Clustering coefficient:} We use the clustering coefficient for every vertex as a feature. We first define this coefficient for unweighted graphs. We denote the number of triangles through vertex $u$ as $T(u)$. Then the clustering coefficient $c_u$ of vertex $u$ in an unweighted graph is the fraction of possible triangles through $u$: \begin{equation}
        c_u = \frac{2T(u)}{deg(u)(deg(u)-1)}
    \end{equation} For weighted graphs, one can define the clustering coefficient in multiple ways~\cite{saramaki2007generalizations}. We rely on the method that computes the geometric average of the subgraph edge weights~\cite{onnela2005intensity}: \begin{equation}
        c_u = \frac{1}{deg(u)(deg(u)-1)}\sum_{vw}(\expandafter\hat{w}_{uv}\expandafter\hat{w}_{uw}\expandafter\hat{w}_{vw})^{1/3}
    \end{equation} Here $\expandafter\hat{w}_{uv}$ are normalized edge weights, $\expandafter\hat{w}_{uv} = w_{uv}/max(w)$, where $max(w)$ is the maximum edge weight in the network.
\end{enumerate}

\begin{figure*}
\minipage{0.23\textwidth}
  \includegraphics[width=\linewidth]{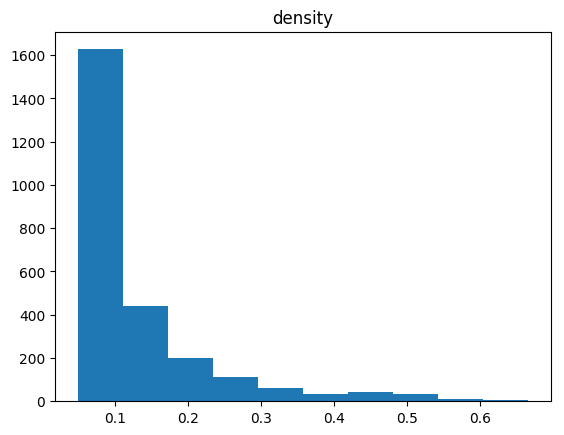}
  \subcaption{Erd\H{o}s--R\'{e}nyi (ER) model}
\endminipage\hfill
\minipage{0.23\textwidth}
  \includegraphics[width=\linewidth]{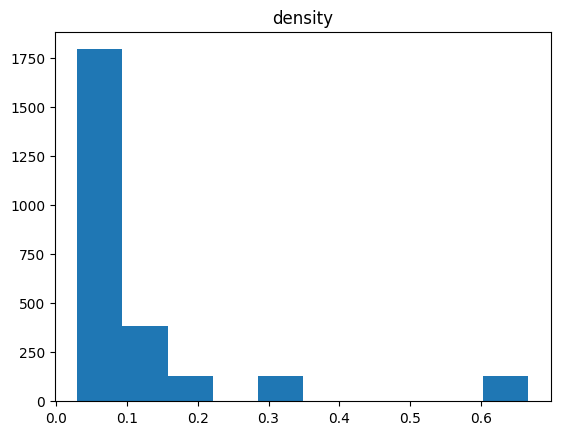}
  \subcaption{Watts-Strogatz (WS) model}
\endminipage\hfill
\minipage{0.23\textwidth}%
  \includegraphics[width=\linewidth]{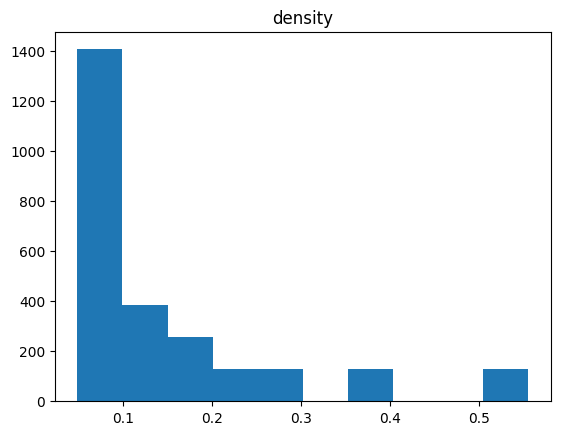}
  \subcaption{Barab\'{a}si--Albert (BA) model}
\endminipage\hfill
\minipage{0.23\textwidth}%
  \includegraphics[width=\linewidth]{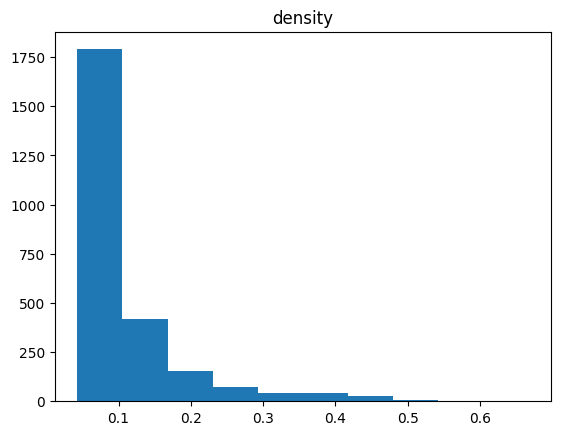}
  \subcaption{Random geometric graph (GE) model}
\endminipage
\caption{Density distribution}
\label{fig:density}
\end{figure*}

\begin{figure*}
\minipage{0.23\textwidth}
  \includegraphics[width=\linewidth]{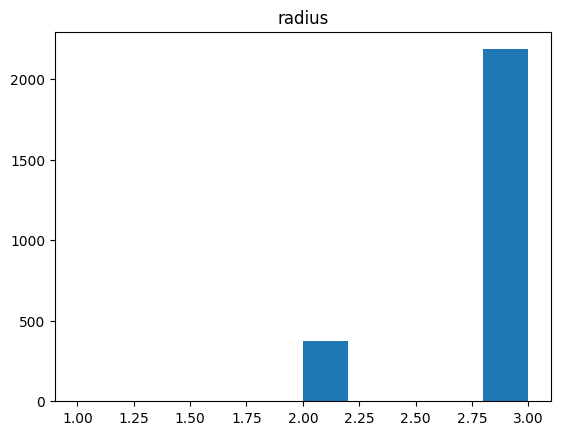}
  \subcaption{Erd\H{o}s--R\'{e}nyi (ER) model}
\endminipage\hfill
\minipage{0.23\textwidth}
  \includegraphics[width=\linewidth]{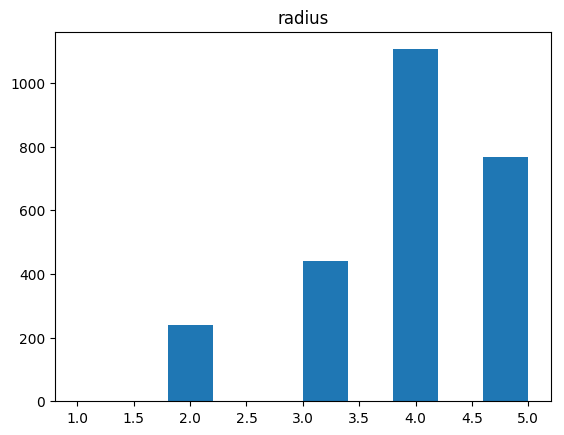}
  \subcaption{Watts-Strogatz (WS) model}
\endminipage\hfill
\minipage{0.23\textwidth}%
  \includegraphics[width=\linewidth]{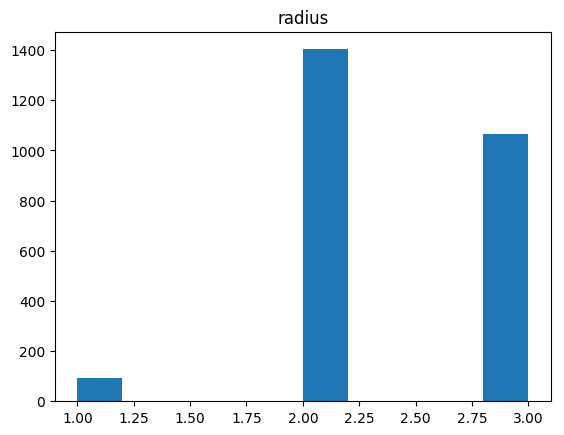}
  \subcaption{Barab\'{a}si--Albert (BA) model}
\endminipage\hfill
\minipage{0.23\textwidth}%
  \includegraphics[width=\linewidth]{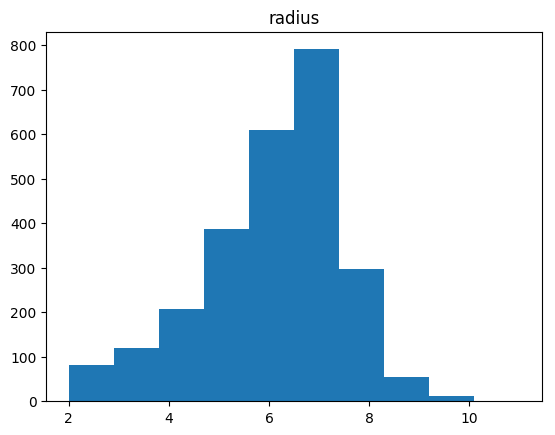}
  \subcaption{Random geometric graph (GE) model}
\endminipage
\caption{Radius distribution}
\label{fig:radius}
\end{figure*}

\section{Experimental results and analyses}\label{sec:experiment}

Before proceeding with the learning experiments, we note some characeristics of the randomly generated graphs -- namely the distributions of the density (Figure \ref{fig:density}) and radius (Figure \ref{fig:radius}) of the graphs from each generator.

%We have considered four random graph generation models. Different models have different densities and radius distributions. For example, we show the density and radius distribution of a set of graphs in Figure~\ref{fig:density} and Figure~\ref{fig:radius} respectively. 
Geometric graphs have much larger radii on average than the others, as the other generators create high degree nodes that make the radius of the graphs relatively small. The presence of many high degree nodes makes the Steiner tree problem simpler, as terminal nodes are closely connected more often, and there is often no need to include other nodes in the solution. This also reflects the learning process of the neural network models: after 500 epochs, the loss decreases significantly for datasets generated from only one of the Erd\H{o}s--R\'{e}nyi, Watts-Strogatz, and Barab\'{a}si--Albert models, while the training for datasets generated solely from the geometric model typically requires around 1000 epochs to significantly reduce the loss. Hence, in the combined dataset we train the models for 1000 epochs.

We have used $80\%$ of the problem instances for training and the rest for testing.  Table~\ref{tab:unweigthed_max} shows the results of testing on unweighted graphs. 
As a baseline for comparison with standard algorithms, we compare our learning results with the 2-approximation algorithm described above. The Steiner tree problem is well-studied and there are many different approximation algorithms, see~\cite{hauptmann2013compendium,promel2012steiner,winter1987steiner}, but out of these we choose to compare with the $2$-approximation algorithm given that it has a straightforward implementation and performs very well in practice. In the tables below, FF1, GNN1, GCN1, and GAT1 represent the learning models combined with the simple heuristic of Section \ref{sec:heuristic}, and FF2, GNN2, GCN2, and GAT2 represent the learning models combined with the more advanced heuristic. The numbers in the table represent the experimental approximation ratios: the ratio of the cost of the heuristic solution to the cost of the optimal solution. We can see that the simple heuristic performs relatively poorly. However, it is 5-10 times faster compared to other heuristics (Table~\ref{tab:unweigthed_time}). The Erd\H{o}s--R\'{e}nyi graphs have the largest maximum ratio. The Watts-Strogatz graphs have a relatively larger average ratio. Boldface numbers indicate the best ratio along each column. We can see that most of the time and for most generators, the second heuristic provides the best average experimental approximation ratio. However, due to the complexity of the second heuristic, some of the learning models using it take somewhat longer to run than the 2-approximation. 

\begin{table*}
\centering
\begin{tabular}{|l|l|l|l|l|l|l|l|l|}\hline
Gen. & \multicolumn{2}{|c|}{ER} & \multicolumn{2}{|c|}{WS} & \multicolumn{2}{|c|}{BA} & \multicolumn{2}{|c|}{GE} \\ \hline
Model & Max & Avg & Max & Avg & Max & Avg & Max & Avg \\ \hline
2-approx & 1.4 & 1.04 & 1.28 & 1.05 & 1.6 & 1.006 & 1.28 & \textbf{1.01} \\ \hline
FF1 & 9.0 & 1.59 & 9.0 & 1.37 & 5.8 & 1.4 & 9.0 & 1.73 \\ \hline
GNN1 & 4.54 & 1.36 & 3.71 & 1.15 & 1.5 & 1.01 & 3.94 & 1.68 \\ \hline
GCN1 & 4.25 & 1.35 & 2.42 & 1.13 & 3.0 & 1.04 & 4.14 & 1.66 \\ \hline
GAT1 & 4.25 & 1.35 & 2.43 & 1.13 & 3.0 & 1.04 & 4.12 & 1.65 \\ \hline
FF2 & 1.4 & \textbf{1.03} & 1.23 & \textbf{1.04} & 1.51 & \textbf{1.005} & 1.28 & \textbf{1.01} \\ \hline
GNN2 & 1.33 & \textbf{1.03} & \textbf{1.21} & \textbf{1.04} & \textbf{1.45} & \textbf{1.005} & \textbf{1.2} & \textbf{1.01} \\ \hline
GCN2 & \textbf{1.3} & \textbf{1.03} & 1.23 & \textbf{1.04} & 1.48 & \textbf{1.005} & 1.25 & \textbf{1.01} \\ \hline
GAT2 & 1.33 & \textbf{1.03} & 1.23 & \textbf{1.04} & 1.47 & \textbf{1.005} & 1.26 & \textbf{1.01} \\ \hline
\end{tabular}
\caption{Approximation ratio for unweighted graphs.}\label{tab:unweigthed_max}
\end{table*}

\begin{table*}
\centering
\begin{tabular}{|l|l|l|l|l|}\hline
Mod./Gen. & ER & WS & BA & GE \\ \hline
2-approx & 2484.96 & 1480.52 & 2404.60 & 2014.82 \\ \hline
FF1 & \textbf{351.91} & 286.31 & \textbf{471.58} & 341.14 \\ \hline
GNN1 & 767.80 & 685.19 & 1006.67 & 763.62 \\ \hline
GCN1 & 385.24 & \textbf{188.68} & 741.51 & \textbf{280.95} \\ \hline
GAT1 & 671.86 & 237.48 & 789.47 & 637.85 \\ \hline
FF2 & 3652.59 & 2008.85 & 2676.14 & 2193.26 \\ \hline
GNN2 & 5929.94 & 3641.42 & 5826.08 & 4852.91 \\ \hline
GCN2 & 3048.39 & 2898.11 & 2922.69 & 3713.09 \\ \hline
GAT2 & 5669.98 & 2318.34 & 3656.26 & 4249.00 \\ \hline
\end{tabular}
\caption{Running time (in seconds) for unweighted graphs.}\label{tab:unweigthed_time}
\end{table*}

Results for weighted graphs are shown in Table~\ref{tab:weigthed_max}. For the random graph generators, we draw edge weights uniformly at random from {\color{blue} $\{1, 2, \cdots, 10\}$}. We also use the SteinLib library dataset for testing, as it provides hard instances of the Steiner tree problem. As can be seen from the table, the SteinLib instances have the largest maximum and average ratios, which is to be expected. The Watts-Strogatz model has a relatively lower ratio. Again we compare the different neural network-based models with the $2$-approximation algorithm. The simple heuristics (FF1, GNN1, GCN1, and GAT1) provide a relatively larger ratio, though the running time of these methods is several magnitudes faster compared to the $2$-approximation algorithm. The best ratio in each column is marked in bold. For most of the graph generators, GNN2, GCN2, and GAT2 perform better than the other heuristics on average at the cost of somewhat more computation time.

\begin{table*}
\centering
\begin{tabular}{|l|l|l|l|l|l|l|l|l|l|l|}\hline
Gen. & \multicolumn{2}{|c|}{ER} & \multicolumn{2}{|c|}{WS} & \multicolumn{2}{|c|}{BA} & \multicolumn{2}{|c|}{GE} & \multicolumn{2}{|c|}{SL} \\ \hline
Model & Max & Avg & Max & Avg & Max & Avg & Max & Avg & Max & Avg \\ \hline
2-approx & 1.21 & 1.03 & 1.15 & \textbf{1.001} & 1.25 & 1.008 & 1.24 & 1.03 & 1.45 & 1.24 \\ \hline
FF1 & 8.77 & 1.67 & 9.08 & 1.39 & 4.62 & 1.39 & 10.30 & 1.87 & 10.29 & 2.23 \\ \hline
GNN1 & 4.01 & 1.43 & 3.43 & 1.18 & 1.26 & 1.12 & 3.91 & 1.81 & 4.56 & 2.15 \\ \hline
GCN1 & 3.76 & 1.43 & 2.26 & 1.17 & 2.43 & 1.13 & 4.1 & 1.79 & 4.78 & 2.13 \\ \hline
GAT1 & 3.76 & 1.43 & 2.27 & 1.17 & 2.43 & 1.14 & 3.29 & 1.78 & 3.83 & 2.14 \\ \hline
FF2 &  1.2 & 1.03 & 1.13 & \textbf{1.001} & 1.25 & 1.007 & 1.24 & 1.03 & 1.45 & 1.23 \\ \hline
GNN2 & \textbf{1.19} & \textbf{1.02} & \textbf{1.12} & \textbf{1.001} & \textbf{1.23} & \textbf{1.006} & 1.23 & \textbf{1.02} & \textbf{1.43} & \textbf{1.21} \\ \hline
GCN2 & 1.2 & \textbf{1.02} & 1.13 & \textbf{1.001} & 1.24 & \textbf{1.006} & \textbf{1.22} & \textbf{1.02} & 1.44 & \textbf{1.21} \\ \hline
GAT2 & \textbf{1.19} & \textbf{1.02} & 1.13 & \textbf{1.001} & \textbf{1.23} & \textbf{1.006} & \textbf{1.22} & \textbf{1.02} & \textbf{1.43} & \textbf{1.21} \\ \hline
\end{tabular}
\caption{Approximation ratio for weighted graphs.}\label{tab:weigthed_max}
\end{table*}

\begin{table*}
\centering
\begin{tabular}{|l|l|l|l|l|l|}\hline
Mod./Gen. & ER & WS & BA & GE & SL \\ \hline
2-approx & 624.09 & 350.64 & 628.72 & 511.73 & 154.91 \\ \hline
FF1 & \textbf{46.84} & \textbf{32.14} & \textbf{48.56} & \textbf{44.45} & 13.06 \\ \hline
GNN1 & 72.47 & 35.28 & 91.00 & 95.48 & 19.74 \\ \hline
GCN1 & 52.99 & 32.65 & 54.77 & 51.64 & \textbf{12.62} \\ \hline
GAT1 & 85.52 & 33.45 & 71.65 & 69.49 & 15.74 \\ \hline
FF2 & 719.30 & 384.06 & 1146.40 & 969.99 & 266.22 \\ \hline
GNN2 & 1491.36 & 606.85 & 1273.60 & 842.73 & 248.34 \\ \hline
GCN2 & 1245.13 & 582.06 & 1026.96 & 795.15 & 223.80 \\ \hline
GAT2 & 1216.86 & 644.32 & 1102.08 & 992.39 & 260.77 \\ \hline
\end{tabular}
\caption{Running time (in seconds) for weighted graphs.}\label{tab:weigthed_time}
\end{table*}

\section{Conclusion}

We have used different neural network models to compute Steiner trees for weighted and unweighted graphs, and compared the output to the exact solutions. We trained the models on a combination of randomly generated graphs from various models. For testing on weighted graphs, we also added SteinLib instances. The models here output a binary vector indicating whether a given node is part of the Steiner tree or not, and therefore we must determine a rule to include or exclude edges in the final Steiner tree solution. To this end, we first proposed a simple heuristic that can compute the Steiner tree faster than the classical $2$-approximation algorithm; however, the experimental approximation ratio of this heuristic worse than the $2$-approximation in practice. We then proposed another heuristic that uses more computation but performs better on average than the $2$-approximation. Our study shows that the application of different combinatorial techniques along with neural network prediction can provide better solutions to the Steiner tree problem. 

For future work, we believe that the application of other combinatorial techniques as a post-processing step as well as other models like reinforcement learning remain an interesting avenue for exploration. Additionally, more sophisticated GNN models which output edge information should be explored, as this could potentially eliminate the need for running a heuristic after the output of the learning model.  Finally, some GNN models indicate that the use of additional input information, such as a candidate solution to the Steiner tree problem, into the learning model can yield much better success. Applying this technique to similar models than those presented here could lead to significantly better learning models.

\bibliographystyle{plain}
\bibliography{references}

\begin{thebibliography}{10}

\bibitem{agrawal1995trees}
Ajit Agrawal, Philip Klein, and Ramamoorthi Ravi.
\newblock When trees collide: An approximation algorithm for the generalized
  steiner problem on networks.
\newblock {\em SIAM journal on Computing}, 24(3):440--456, 1995.

\bibitem{ahmed2019multi}
Reyan Ahmed, Patrizio Angelini, Faryad~Darabi Sahneh, Alon Efrat, David
  Glickenstein, Martin Gronemann, Niklas Heinsohn, Stephen~G Kobourov, Richard
  Spence, Joseph Watkins, et~al.
\newblock Multi-level steiner trees.
\newblock {\em Journal of Experimental Algorithmics (JEA)}, 24:1--22, 2019.

\bibitem{barabasi1999emergence}
Albert-L{\'a}szl{\'o} Barab{\'a}si and R{\'e}ka Albert.
\newblock Emergence of scaling in random networks.
\newblock {\em science}, 286(5439):509--512, 1999.

\bibitem{battaglia2016interaction}
Peter Battaglia, Razvan Pascanu, Matthew Lai, Danilo~Jimenez Rezende, and Koray
  Kavukcuoglu.
\newblock Interaction networks for learning about objects, relations and
  physics.
\newblock In {\em Advances in neural information processing systems}, pages
  4502--4510, 2016.

\bibitem{bello2016neural}
Irwan Bello, Hieu Pham, Quoc~V Le, Mohammad Norouzi, and Samy Bengio.
\newblock Neural combinatorial optimization with reinforcement learning.
\newblock In {\em Proceedings of The International Conference on Learning
  Representations (ICLR)}, 2016.

\bibitem{cormen2009introduction}
Thomas~H Cormen, Charles~E Leiserson, Ronald~L Rivest, and Clifford Stein.
\newblock {\em Introduction to algorithms}.
\newblock MIT press, 2009.

\bibitem{dai2020framework}
Hanjun Dai, Xinshi Chen, Yu~Li, Xin Gao, and Le~Song.
\newblock A framework for differentiable discovery of graph algorithms.
\newblock In {\em Proceedings of Neural Information Processing Systems}, 2020.

\bibitem{dai2017learning}
Hanjun Dai, Elias~B Khalil, Yuyu Zhang, Bistra Dilkina, and Le~Song.
\newblock Learning combinatorial optimization algorithms over graphs.
\newblock In {\em Proceedings of Neural Information Processing Systems}, 2017.

\bibitem{erdos1959random}
Paul Erd{\H o}s and Alfr{\'e}d R{\'e}nyi.
\newblock On random graphs, i.
\newblock {\em Publicationes Mathematicae (Debrecen)}, 6:290--297, 1959.

\bibitem{fout2017protein}
Alex Fout, Jonathon Byrd, Basir Shariat, and Asa Ben-Hur.
\newblock Protein interface prediction using graph convolutional networks.
\newblock In {\em Advances in neural information processing systems}, pages
  6530--6539, 2017.

\bibitem{garey1979computers}
Michael~R Garey and David~S Johnson.
\newblock {\em Computers and intractability}, volume 174.
\newblock freeman San Francisco, 1979.

\bibitem{garg2020generalization}
Vikas Garg, Stefanie Jegelka, and Tommi Jaakkola.
\newblock Generalization and representational limits of graph neural networks.
\newblock In {\em International Conference on Machine Learning}, pages
  3419--3430, 2020.

\bibitem{gilmer2017neural}
Justin Gilmer, Samuel~S Schoenholz, Patrick~F Riley, Oriol Vinyals, and
  George~E Dahl.
\newblock Neural message passing for quantum chemistry.
\newblock In {\em International conference on machine learning}, pages
  1263--1272, 2017.

\bibitem{gori2005new}
Marco Gori, Gabriele Monfardini, and Franco Scarselli.
\newblock A new model for learning in graph domains.
\newblock In {\em Proceedings. 2005 IEEE International Joint Conference on
  Neural Networks, 2005.}, volume~2, pages 729--734. IEEE, 2005.

\bibitem{hamaguchi2017knowledge}
Takuo Hamaguchi, Hidekazu Oiwa, Masashi Shimbo, and Yuji Matsumoto.
\newblock Knowledge transfer for out-of-knowledge-base entities: A graph neural
  network approach.
\newblock In {\em Proceedings of the 26th International Joint Conference on
  Artificial Intelligence}, 2017.

\bibitem{hamilton2017inductive}
Will Hamilton, Zhitao Ying, and Jure Leskovec.
\newblock Inductive representation learning on large graphs.
\newblock In {\em Advances in neural information processing systems}, pages
  1024--1034, 2017.

\bibitem{HAMMOND2011129}
David~K. Hammond, Pierre Vandergheynst, and Rémi Gribonval.
\newblock Wavelets on graphs via spectral graph theory.
\newblock {\em Applied and Computational Harmonic Analysis}, 30(2):129--150,
  2011.

\bibitem{hauptmann2013compendium}
Mathias Hauptmann and Marek Karpi{\'n}ski.
\newblock {\em A compendium on Steiner tree problems}.
\newblock Inst. f{\"u}r Informatik, 2013.

\bibitem{karp1972reducibility}
Richard~M Karp.
\newblock Reducibility among combinatorial problems.
\newblock In {\em Complexity of computer computations}, pages 85--103.
  Springer, 1972.

\bibitem{khalil2017learning}
Elias Khalil, Hanjun Dai, Yuyu Zhang, Bistra Dilkina, and Le~Song.
\newblock Learning combinatorial optimization algorithms over graphs.
\newblock In {\em Advances in Neural Information Processing Systems}, pages
  6348--6358, 2017.

\bibitem{kingma2014adam}
Diederik~P Kingma and Jimmy Ba.
\newblock Adam: A method for stochastic optimization.
\newblock {\em arXiv preprint arXiv:1412.6980}, 2014.

\bibitem{kipf2016semi}
Thomas~N Kipf and Max Welling.
\newblock Semi-supervised classification with graph convolutional networks.
\newblock In {\em Proceedings of The International Conference on Learning
  Representations (ICLR)}, 2016.

\bibitem{KMV00}
T.~Koch, A.~Martin, and S.~{Vo\ss}.
\newblock {SteinLib}: An updated library on {Steiner} tree problems in graphs.
\newblock Technical Report ZIB-Report 00-37, Konrad-Zuse-Zentrum {f\"ur}
  Informationstechnik Berlin, Takustr. 7, Berlin, 2000.

\bibitem{kool2018attention}
Wouter Kool, Herke Van~Hoof, and Max Welling.
\newblock Attention, learn to solve routing problems!
\newblock In {\em Proceedings of The International Conference on Learning
  Representations (ICLR)}, 2018.

\bibitem{lei2017deriving}
Tao Lei, Wengong Jin, Regina Barzilay, and Tommi Jaakkola.
\newblock Deriving neural architectures from sequence and graph kernels.
\newblock In {\em International Conference on Machine Learning}, pages
  2024--2033, 2017.

\bibitem{lemos2019graph}
Henrique Lemos, Marcelo Prates, Pedro Avelar, and Luis Lamb.
\newblock Graph colouring meets deep learning: Effective graph neural network
  models for combinatorial problems.
\newblock In {\em 2019 IEEE 31st International Conference on Tools with
  Artificial Intelligence (ICTAI)}, pages 879--885. IEEE, 2019.

\bibitem{li2018combinatorial}
Zhuwen Li, Qifeng Chen, and Vladlen Koltun.
\newblock Combinatorial optimization with graph convolutional networks and
  guided tree search.
\newblock In {\em Proceedings of Neural Information Processing Systems}, 2018.

\bibitem{mishra2020node}
Pushkar Mishra, Aleksandra Piktus, Gerard Goossen, and Fabrizio Silvestri.
\newblock Node masking: Making graph neural networks generalize and scale
  better.
\newblock {\em arXiv preprint arXiv:2001.07524}, 2020.

\bibitem{narayanan2017graph2vec}
Annamalai Narayanan, Mahinthan Chandramohan, Rajasekar Venkatesan, Lihui Chen,
  Yang Liu, and Shantanu Jaiswal.
\newblock graph2vec: Learning distributed representations of graphs.
\newblock {\em arXiv preprint arXiv:1707.05005}, 2017.

\bibitem{onnela2005intensity}
Jukka-Pekka Onnela, Jari Saram{\"a}ki, J{\'a}nos Kert{\'e}sz, and Kimmo Kaski.
\newblock Intensity and coherence of motifs in weighted complex networks.
\newblock {\em Physical Review E}, 71(6):065103, 2005.

\bibitem{penrose2003random}
Mathew Penrose.
\newblock {\em Random geometric graphs}.
\newblock Number~5. Oxford university press, 2003.

\bibitem{prates2019learning}
Marcelo Prates, Pedro~HC Avelar, Henrique Lemos, Luis~C Lamb, and Moshe~Y
  Vardi.
\newblock Learning to solve np-complete problems: A graph neural network for
  decision tsp.
\newblock In {\em Proceedings of the AAAI Conference on Artificial
  Intelligence}, volume~33, pages 4731--4738, 2019.

\bibitem{promel2012steiner}
Hans~J{\"u}rgen Pr{\"o}mel and Angelika Steger.
\newblock {\em The Steiner tree problem: a tour through graphs, algorithms, and
  complexity}.
\newblock Springer Science \& Business Media, 2012.

\bibitem{sanchez2018graph}
Alvaro Sanchez-Gonzalez, Nicolas Heess, Jost~Tobias Springenberg, Josh Merel,
  Martin Riedmiller, Raia Hadsell, and Peter Battaglia.
\newblock Graph networks as learnable physics engines for inference and
  control.
\newblock In {\em International Conference on Machine Learning}, 2018.

\bibitem{saramaki2007generalizations}
Jari Saram{\"a}ki, Mikko Kivel{\"a}, Jukka-Pekka Onnela, Kimmo Kaski, and Janos
  Kertesz.
\newblock Generalizations of the clustering coefficient to weighted complex
  networks.
\newblock {\em Physical Review E}, 75(2):027105, 2007.

\bibitem{scarselli2008graph}
Franco Scarselli, Marco Gori, Ah~Chung Tsoi, Markus Hagenbuchner, and Gabriele
  Monfardini.
\newblock The graph neural network model.
\newblock {\em IEEE Transactions on Neural Networks}, 20(1):61--80, 2008.

\bibitem{scarselli1998universal}
Franco Scarselli and Ah~Chung Tsoi.
\newblock Universal approximation using feedforward neural networks: A survey
  of some existing methods, and some new results.
\newblock {\em Neural networks}, 11(1):15--37, 1998.

\bibitem{selsam2018learning}
Daniel Selsam, Matthew Lamm, Benedikt B{\"u}nz, Percy Liang, Leonardo de~Moura,
  and David~L Dill.
\newblock Learning a sat solver from single-bit supervision.
\newblock In {\em Proceedings of The International Conference on Learning
  Representations (ICLR)}, 2018.

\bibitem{shuman2016vertex}
David~I Shuman, Benjamin Ricaud, and Pierre Vandergheynst.
\newblock Vertex-frequency analysis on graphs.
\newblock {\em Applied and Computational Harmonic Analysis}, 40(2):260--291,
  2016.

\bibitem{velickovic2018graph}
Petar Veli{\v{c}}kovi{\'{c}}, Guillem Cucurull, Arantxa Casanova, Adriana
  Romero, Pietro Li{\`{o}}, and Yoshua Bengio.
\newblock {Graph Attention Networks}.
\newblock {\em International Conference on Learning Representations}, 2018.

\bibitem{vesselinova2020learning}
Natalia Vesselinova, Rebecca Steinert, Daniel~F. Perez-Ramirez, and Magnus
  Boman.
\newblock Learning combinatorial optimization on graphs: A survey with
  applications to networking.
\newblock {\em IEEE Access}, 8:120388--120416, 2020.

\bibitem{watts1998collective}
Duncan~J Watts and Steven~H Strogatz.
\newblock Collective dynamics of ‘small-world’networks.
\newblock {\em Nature}, 393(6684):440, 1998.

\bibitem{winter1987steiner}
Pawel Winter.
\newblock Steiner problem in networks: A survey.
\newblock {\em Networks}, 17(2):129--167, 1987.

\bibitem{ying2019gnnexplainer}
Rex Ying, Dylan Bourgeois, Jiaxuan You, Marinka Zitnik, and Jure Leskovec.
\newblock Gnnexplainer: Generating explanations for graph neural networks.
\newblock {\em Advances in neural information processing systems}, 32:9240,
  2019.

\bibitem{zhou2018graph}
Jie Zhou, Ganqu Cui, Shengding Hu, Zhengyan Zhang, Cheng Yang, Zhiyuan Liu,
  Lifeng Wang, Changcheng Li, and Maosong Sun.
\newblock Graph neural networks: A review of methods and applications.
\newblock {\em AI Open}, 1:57--81, 2020.

\end{thebibliography}
\end{document}